\documentclass[journal,twoside,web]{ieeecolor}
\usepackage{generic}
\usepackage{cite}
\usepackage{amsmath,amssymb,amsfonts}
\usepackage{algorithmic}
\usepackage{graphicx}
\usepackage{algorithm,algorithmic}
\usepackage{hyperref}
\hypersetup{hidelinks=true}
\usepackage{textcomp}
\usepackage{booktabs}
\usepackage{array}
\usepackage{multirow}

\def\BibTeX{{\rm B\kern-.05em{\sc i\kern-.025em b}\kern-.08em
    T\kern-.1667em\lower.7ex\hbox{E}\kern-.125emX}}
\begin{document}
\title{An Interpretable Multi-Plane Fusion Framework With Kolmogorov–Arnold Network Guided Attention Enhancement for Alzheimer’s Disease Diagnosis}

\author{Xiaoxiao Yang, Meiliang Liu, \IEEEmembership{Member, IEEE}, Yunfang Xu, Zijin Li, Zhengye Si, Xinyue Yang, and Zhiwen Zhao
\thanks{This work has been granted by National Natural Science Foundation of China under Grant 61075075. The work is also supported by the Research Center for Intelligent Engineering and Educational Application of Beijing Normal University at Zhuhai campus. \textit{(Corresponding author: Zhiwen Zhao.)}}
\thanks{Xiaoxiao Yang, Meiliang Liu, Yunfang Xu, Zijin Li, Zhengye Si, and Xinyue Yang are with the School of Artificial Intelligence, Beijing Normal University, Beijing, China (e-mail: yangxiaoxiao@mail.bnu.edu.cn; liumeiliang520@gmail.com; xuyunfang@mail.bnu.edu.cn; 2291832685@qq.com; sizhengye0302@gmail.com; 3215253513@qq.com).}
\thanks{Zhiwen Zhao is with the School of Artificial Intelligence, Beijing Normal University, Beijing, China, also with the Advanced Institute of Natural Sciences, Beijing Normal University, Zhuhai, Guangdong, China (e-mail:mlt.bnu2017@bnu.edu.cn).}
}

\maketitle

\begin{abstract}
Alzheimer’s disease (AD) is a progressive neurodegenerative disorder that severely impairs cognitive function and quality of life. Timely intervention in AD relies heavily on early and precise diagnosis, which remains challenging due to the complex and subtle structural changes in the brain. Most existing deep learning methods focus only on a single plane of structural magnetic resonance imaging (sMRI) and struggle to accurately capture the complex and nonlinear relationships among pathological regions of the brain, thus limiting their ability to precisely identify atrophic features. To overcome these limitations, we propose an innovative framework, MPF-KANSC, which integrates multi-plane fusion (MPF) for combining features from the coronal, sagittal, and axial planes, and a Kolmogorov-Arnold Network-guided spatial-channel attention mechanism (KANSC) to more effectively learn and represent sMRI atrophy features. Specifically, the proposed model enables parallel feature extraction from multiple anatomical planes, thus capturing more comprehensive structural information. The KANSC attention mechanism further leverages a more flexible and accurate nonlinear function approximation technique, facilitating precise identification and localization of disease-related abnormalities. Experiments on the ADNI dataset confirm that the proposed MPF-KANSC achieves superior performance in AD diagnosis. Moreover, our findings provide new evidence of right-lateralized asymmetry in subcortical structural changes during AD progression, highlighting the model’s promising interpretability.
\end{abstract}


\begin{figure*}
	\centering
	\includegraphics[width=\textwidth]{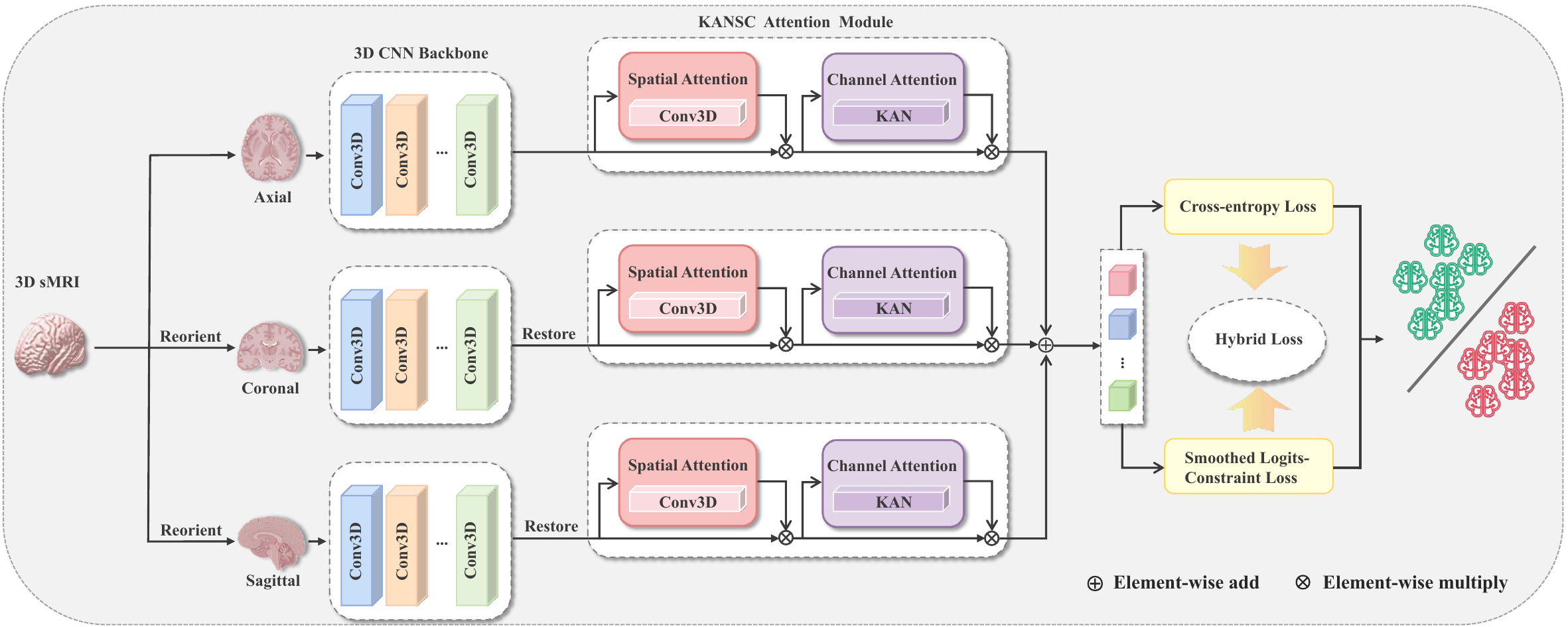}
	\caption{Overall architecture of the proposed MPF-KANSC framework. The model consists of three parallel 3D CNN branches for multi-plane feature extraction, each with a KANSC attention module to enhance spatial and channel features. The outputs are fused and optimized by a hybrid loss module that combines the cross-entropy loss and the smoothed logits-constraint loss.}
	\label{FIG:01}
\end{figure*}

\section{Introduction}
\label{introduction}
\IEEEPARstart{A}{s} the most common cause of dementia, Alzheimer’s disease (AD) is primarily characterized by progressive memory impairment, which can advance to severe cognitive decline and neuropsychiatric symptoms, profoundly affecting individuals, families, and society \cite{zhang2024recent}. The prodromal stage, mild cognitive impairment (MCI), includes progressive MCI (pMCI), which often develops into AD within 36 months, and stable MCI (sMCI), which remains unchanged \cite{thung2018conversion}. Early identification of MCI is crucial, as disease-modifying therapies are most effective at this stage \cite{garcia2009existing}, \cite{he2019amyloid}. Structural magnetic resonance imaging (sMRI) offers a high-resolution, non-invasive tool for AD diagnosis \cite{lombardi2020structural}, \cite{chouliaras2023use}, but traditional assessment is limited by reliance on radiologist interpretation, which is time-consuming and variable. Consequently, deep learning-based automated diagnostic methods have been developed to enhance efficiency and accuracy and address the shortage of AD specialists.

In recent years, deep neural networks have attracted considerable attention for their effectiveness in extracting discriminative features from sMRI. According to the input type of sMRI \cite{wen2020convolutional}, existing deep learning models can be classified into the following categories: 2D slice-level, 3D patch-level, region of interest (ROI), and 3D subject-level approaches. 2D slice-level techniques typically select a limited number of representative slices from the sagittal plane \cite{francis2023ensemble}, \cite{hoang2023vision} or the axial plane \cite{hu2023vgg} for diagnosis. While these methods provide detailed local information, they fail to capture the global context of the entire brain, which may reduce diagnostic accuracy. 3D patch-level strategies recover local volumetric information by extracting several subvolumes \cite{zhang2023smri}, \cite{qiang2025classification}, but typically require independent network training for each patch and are unable to model interactions between patches. ROI-based models focus on specific brain regions \cite{feng2022detection}, \cite{lei2024hybrid}, like the hippocampus, to minimize noise, but this limits the analysis to a few predefined areas and may overlook disease-related changes elsewhere in the brain. 3D subject-level methods utilize the entire brain \cite{pei2022multi}, \cite{tong2024research}, \cite{xu2024interpretable} and capture comprehensive anatomical information, but often struggle to accurately localize focal pathological regions in patients.

To more adaptively learn both global and local discriminative features from sMRI, recent studies have introduced attention mechanisms into deep learning models. By enhancing salient information and suppressing irrelevant signals, these methods have achieved promising results in improving diagnostic performance \cite{jin2019attention}, \cite{wu2022attention}, \cite{xu2024interpretable}. Nevertheless, many current models focus on feature extraction from a single plane and therefore cannot fully leverage the complementary information available across axial, coronal, and sagittal planes. A few approaches have attempted to extract features from multi-plane data using conventional attention modules \cite{liu2023mps}, \cite{wei20253d}, \cite{rahim2025early}, but the fixed activation functions in these modules limit their ability to capture the complex and nonlinear patterns within each anatomical plane. This limitation highlights the need for more flexible architectures capable of learning intricate disease characteristics. Kolmogorov–Arnold Network (KAN) provides an innovative solution by replacing static weight parameters at the network’s edges with adaptive spline-parameterized functions \cite{liu2024kan}, thereby dynamically adapting to data patterns and approximating complex, non-smooth functions. It is this adaptive capability that enables KAN to demonstrate unique advantages across various scientific applications \cite{ji2024comprehensive}. 

Therefore, we propose a AD diagnosis model based on a KAN-guided attention mechanism that extracts features from the axial, coronal, and sagittal planes of sMRI. Moreover, we employ a hybrid loss function framework to more accurately enhance the weighting of salient regions. Fig. \ref{FIG:01} presents the overall architecture of the proposed model.

The main contributions of this work are outlined below:
\begin{enumerate}
\item We propose a novel and interpretable deep learning model, namely MPF-KANSC, which leverages the whole 3D sMRI to extract key pathological features from multiple planes, enabling accurate end-to-end diagnosis.
\item We design a new KAN-guided spatial-channel attention mechanism that flexibly models complex nonlinear relationships in sMRI data using a unique edge activation scheme, thereby enhancing the model’s learning capacity.
\item We introduce a smoothed logits-constraint loss function, combined with cross-entropy, to improve the balance between global and local pattern modeling.
\item Extensive experimental results demonstrate that our model enables accurate localization of pathological regions with better interpretability and further reveals pronounced right-lateralized asymmetry in subcortical structures during AD progression.
\end{enumerate}

The remainder of this article is structured as follows. Section \ref{related work} presents a concise review of related studies. Section \ref{model architecture} presents our proposed MPF-KANSC framework in detail. Section \ref{experiment} elaborates on the datasets and preprocessing procedures adopted in our experiments, shows and discusses the experimental results. Finally, Section \ref{conclusion} concludes the study and points out the future research directions.

\begin{figure*}[!t]
	\centering
	\includegraphics[width=\textwidth]{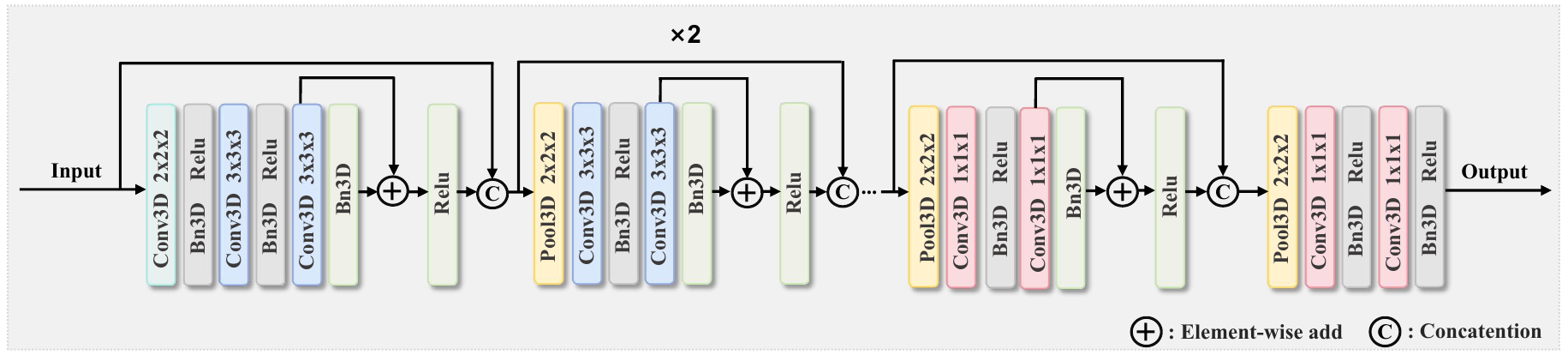}
	\caption{The architecture of 3D CNN backbone. It consists of five hierarchical stages, each composed of convolutional or pooling operations, with repeated use of residual connections via element-wise addition and feature fusion by channel-wise concatenation.}
	\label{FIG:02}
\end{figure*}

\section{Related work}
\label{related work}
In this section, we briefly review some attention-based frameworks for AD diagnosis, the theory and applications of KAN, and recent advances in transfer learning related to this field.

\subsection{Attention-Based Frameworks for AD Diagnosis}
Attention mechanisms have been widely incorporated into deep learning models to enhance their information processing capabilities \cite{niu2021review}. Typically, attention assigns weights to input features and aggregates the most task-relevant information, thereby improving both inference performance and interpretability. In particular, selective attention strategies have been extensively applied in AD analysis to strengthen the model's ability to identify and focus on critical pathological features.

In AD diagnosis based solely on 3D sMRI, researchers have introduced various attention mechanisms to boost model performance \cite{zhang2021explainable}, \cite{luo2023class}, \cite{qin20223d}. For instance, Zhang et al. \cite{zhang2021explainable} proposed ResAttNet, which incorporates a residual self-attention block to capture local, global, and spatial cues, thereby improving diagnostic accuracy. Luo et al. \cite{luo2023class} developed Class Activation Attention Transfer, a novel technique that transfers attention maps from a labeled source classification task to enhance performance on the target task. Additionally, Qin et al. \cite{qin20223d} introduced a hybrid-domain attention mechanism combining channel and spatial attention to differentiate between amnestic MCI and semantic MCI. In the field of multimodal fusion, attention-based designs have also been proven effective \cite{lian2021multi}, \cite{liu2024hammf}. For example, Lian et al. \cite{lian2021multi} presented a Multi-task Weakly-supervised Attention Network that automatically identifies subject-specific discriminative brain regions and integrates MRI with clinical scores for pathological stage prediction. Recently, Liu et al. \cite{liu2024hammf} proposed a Hierarchical Attention-based Multi-task Multi-modal Fusion model, and demonstrated superior performance in clinical score regression.

While the above methods primarily focus on attention mechanisms within a single plane, recent research has begun to investigate the combined advantages of integrating multi-plane feature information with attention mechanisms for AD diagnosis. Liu et al. \cite{liu2023mps} developed an attention-based framework that performs multi-plane and multiscale feature fusion, combining 3D sMRI data with clinical scores to enhance the localization of pathological regions. Rahim et al. \cite{rahim2025early} extracted central 2D slices from the axial, coronal, and sagittal planes, and then employed CNNs equipped with the Convolutional Block Attention Module to filter noise and yield robust feature representations. However, these methods are limited by fixed activation functions and rigid architectures within their attention modules, which restrict their ability to model complex relationships as data complexity increases. Consequently, conventional attention mechanisms can no longer satisfy the evolving requirements of advanced AD diagnosis, highlighting the need for more flexible and innovative approaches.

\subsection{Theory and Applications of KAN}
As a function approximator, Multi-Layer Perceptron (MLP) has long served as a universal backbone in deep learning models \cite{tolstikhin2021mlp}. However, MLPs rely on fixed weight matrices and traditional nonlinear activation functions (such as ReLU or SiLU), which limits their generalization ability when modeling complex visual data. To address these challenges, Liu et al. \cite{liu2024kan} recently introduced the Kolmogorov-Arnold Network (KAN), which is based on the Kolmogorov-Arnold (KA) representation theorem \cite{schmidt2021kolmogorov}, asserting that any multivariate continuous function can be expressed as a finite sum of continuous univariate functions. Specifically, KAN replaces fixed activation functions with spline-parameterized edge activations, providing greater flexibility and accuracy in function approximation. This design effectively alleviates common issues such as gradient vanishing and saturation, enabling the model to capture more complex nonlinear relationships.

Recent studies have shown that KAN possesses outstanding feature extraction capabilities, facilitating its application across a variety of fields \cite{ji2024comprehensive}, including time-series forecasting  \cite{xu2024kolmogorov}, \cite{liu2025kolmogorov}, computer vision \cite{mahara2024dawn}, \cite{ning2025kan}, physics-informed modeling \cite{jacob2024spikans}, \cite{shuai2025physics} and healthcare \cite{tang20243d}, \cite{jahin2024kacq}. In practical scenarios, the unique edge activation mechanism of KAN enhances both modularity and interpretability, which are essential properties for developing advanced AD diagnostic frameworks based on 3D sMRI data. In this study, we develop our MPF-KANSC architecture on the code of efficientKAN (\url{https://github.com/Blealtan/efficient-kan}).

\begin{figure*}[ht]
	\centering
	\includegraphics[width=\textwidth]{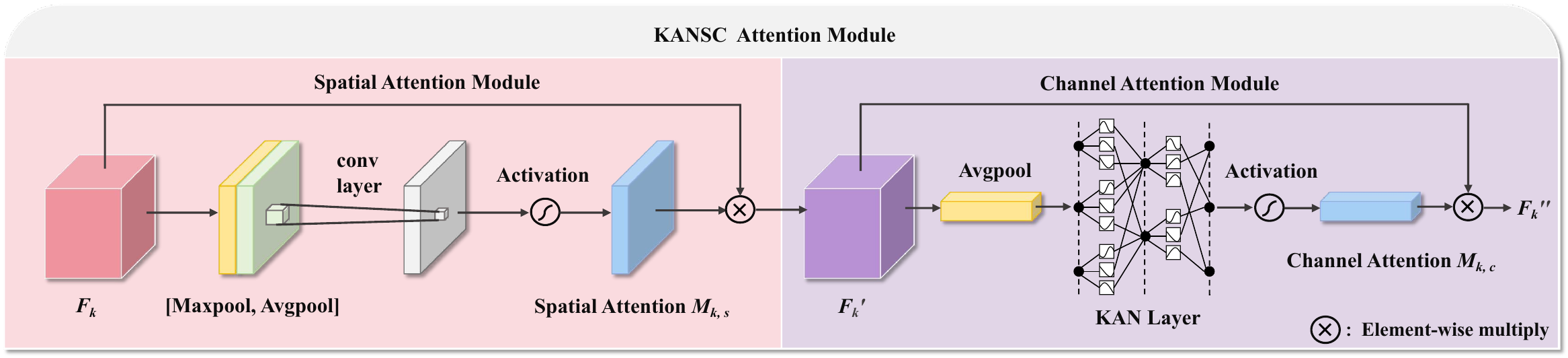}
	\caption{The framework of the KANSC attention module. The left and right panels illustrate the spatial and channel attention submodules, respectively, which together effectively capture discriminative spatial and channel information from the input features.}
	\label{FIG:03}
\end{figure*}

\subsection{Transfer Learning}
Transfer learning is a methodology that enhances the efficacy of a target learner in the target domain by leveraging information obtained from disparate yet connected source domains \cite{zhuang2020comprehensive}, which has been proven effective for early AD diagnosis as well. Compared to the classification task between CN and AD subjects, predicting MCI conversion is more challenging due to the higher structural similarity between sMCI and pMCI brains. Given the relevance between two classification tasks, many studies \cite{pandey2024enhancing} have shown that features learned from CN and AD patients can facilitate more accurate MCI diagnosis. For example, Lian et al. \cite{lian2018hierarchical} proposed a hierarchical fully convolutional network and showed that using parameters pretrained on AD classification tasks significantly improves MCI conversion prediction. Similarly, Gao et al. \cite{gao2020ad} proposed an age-adjusted neural network that retains the entire pretrained model during fine-tuning, thereby transferring knowledge acquired from age prediction to improve MCI prediction. Thus, transfer learning serves as an effective approach for leveraging knowledge from CN and AD classification to improve MCI conversion prediction.

\section{Model architecture}
\label{model architecture}
This section delineates three major parts of our proposed framework, comprising the 3D CNN backbone, the KANSC attention module, and the hybrid loss module.

\subsection{3D CNN Backbone}
Given the inherent structural characteristics of 3D sMRI, convolutional neural networks (CNNs) remain one of the most important architectures for AD diagnosis. To address the challenges of vanishing gradients and to ensure efficient training of deep networks, residual connections \cite{szegedy2017inception} are incorporated to ensure stable network convergence and facilitate the learning of pathological features. In our MPF-KANSC model, as illustrated in Fig. \ref{FIG:01}, each input 3D sMRI is first reoriented into the axial, coronal, and sagittal planes. These views are then processed in parallel by three 3D CNN branches to extract spatial feature representations, with the detailed architecture of each convolutional branch depicted in Fig. \ref{FIG:02}.

Each 3D CNN backbone consists of five sequential stages to enable hierarchical feature extraction. In the first stage, a 3D convolutional layer with a kernel size of 2×2×2, a stride of 2, and no padding performs initial downsampling and outputs 16 channels. This is followed by two 3×3×3 convolutional layers with padding 1, which are connected via a residual link. The residual output is then concatenated with the stage input along the channel dimension to facilitate feature reuse. In the second and third stages, each begins with a max pooling operation using a kernel size of 2×2×2 and a stride of 2, followed by two 3×3×3 convolutional layers, where the number of output channels is set to 32 in the second stage and 64 in the third stage. Both residual and concatenation connections are employed as before. The fourth and fifth stages also start with max pooling, followed by two 1×1×1 convolutional layers. The number of output channels is 128 in the fourth stage and 256 in the fifth stage. The fourth stage retains residual and concatenation operations, while the fifth outputs directly. This architecture progressively extracts multi-scale semantic features from sMRI, enabling effective modeling of complex pathological patterns.

\subsection{KANSC Attention Module}
Capturing both global and local features from whole-brain sMRI remains a significant challenge, as conventional 3D CNN networks tend to focus primarily on local patterns while overlooking inter-regional dependencies. To overcome this limitation, we design a KAN-guided spatial-channel attention mechanism (KANSC). As depicted in Fig. \ref{FIG:01}, features extracted from the axial, coronal, and sagittal views by the 3D CNN backbone are restored to the axial plane and subsequently processed by individual KANSC modules to enhance global contextual understanding. The internal structure of the KANSC attention module is presented in Fig. \ref{FIG:03} and described in detail below.

Traditional spatial-channel attention module consists of two branches \cite{woo2018cbam}: a spatial branch that extracts spatial dependencies via convolution and a channel branch that uses the MLP to model inter-channel relationships. However, MLP relies on fixed nonlinear activation functions, such as ReLU or sigmoid, and thus often struggles to accurately approximate complex mappings. To enhance the capacity of the channel attention branch for capturing complex feature interactions, we replace the conventional MLP with KAN. Within this design, the learnable univariate functions are parameterized using B-splines \cite{somvanshi2024survey}, which are well known for their smoothness, locality, and fine-grained control over functional shapes, thereby enabling dynamic and adaptable nonlinear mappings during training. By integrating this enhanced channel modeling with spatial attention, the proposed KANSC module adaptively captures intricate spatial–channel dependencies, resulting in more discriminative feature representations and markedly improved diagnostic performance.

In KANSC, the spatial attention module follows a standard design. Let $F_k \in \mathbb{R}^{C \times D \times H \times W}$ denote the feature map from the last convolutional layer for plane $k$, where $k \in \{s, c, a\}$ corresponds to the sagittal, coronal, and axial planes, respectively. Here, $C$ is the number of channels, and $D$, $H$, and $W$ represent the depth, height, and width of the feature map. We compute the channel-wise global maximum and average by applying max pooling and average pooling across the spatial dimensions:
\begin{align}
F_{k,max}(d, h, w) = \max_{c \in \{1, \ldots, C\}}{F_k}(c, d, h, w),
\label{ep01}
\end{align}
\begin{align}
F_{k,avg}(d, h, w) = \frac{1}{C}\sum_{c=1}^{C} {F_k}(c, d, h, w), 
\label{ep02}
\end{align}
where \(F_{k,\max}(d, h, w)\) and \(F_{k,\mathrm{avg}}(d, h, w)\) denote the channel-wise maximum and average feature values at voxel location \((d, h, w)\), respectively. Subsequently, the two feature maps are concatenated along the channel dimension, after which a convolutional layer and a sigmoid activation are applied in sequence to produce the spatial attention map:
\begin{align}
M_{k,s} = \sigma \left( f^{3 \times 3 \times 3} \left( [F_{k,max} ; F_{k,avg}] \right) \right)\in(0,1),
\label{ep03}
\end{align}
where $M_{k,s}$ denotes the attention weight at each spatial position of the input feature map, $\sigma$ is the sigmoid activation, $f^{3\times3\times3}(\cdot)$ denotes a convolution operation with a $3\times3\times3$ kernel, stride 1, and padding 1, which outputs a single-channel feature map, $[\ ;\ ]$ denotes concatenation along the channel dimension. Then, the output of the spatial attention module $F'_k$ is calculated as:
\begin{align}
F'_k = M_{k,s} \odot F_k,
\label{ep04}
\end{align}
where $\odot$ indicates element-wise multiplication.

Next, $F'_k$ is fed into the channel attention module. Our ablation experiments indicate that incorporating global max pooling into the channel attention module leads to extreme activations, which can negatively impact model performance. 
Therefore, we apply global average pooling on each channel, $F'_{k,avg}$ is denoted as:
\begin{align}
F'_{k,avg} = \left[ \frac{1}{DHW} \sum_{d=1}^{D} \sum_{h=1}^{H} \sum_{w=1}^{W} F'_k(c, d, h, w) \right]_{c=1}^{C},
\label{ep05}
\end{align}
where $[\cdots]_{c=1}^{C}$ indicates that the operation inside the brackets is performed for each channel $c$, resulting in a $C$-dimensional vector. Then, the resulting vector is fed into the KAN module, followed by a nonlinear activation function:
\begin{align}
M_{k,c} = \sigma\bigl(\mathrm{KAN}(F'_{k,avg})\bigr)\in(0,1),
\label{ep06}
\end{align}
where $M_{k,c}$ denotes the channel attention weight matrix. $\mathrm{KAN}(\cdot)$ refers to a Kolmogorov–Arnold Network composed of three layers with dimensions $C \rightarrow 32 \rightarrow C$, which is defined as follows:
\begin{align}
\mathrm{KAN}(\mathbf{x}) = (\Phi_{L-1} \circ \Phi_{L-2} \circ \cdots \circ \Phi_0)\, \mathbf{x},
\label{ep07}
\end{align}
where $L=3$, corresponding to the input, hidden, and output layers, respectively. Here, we represent the vector of hidden units at layer $l$ as:
\begin{align}
\mathbf{h}^{l} = 
\underbrace{
\begin{pmatrix}
\phi_{l-1,1,1}(\cdot) & \cdots & \phi_{l-1,1,n_{l-1}}(\cdot) \\
\phi_{l-1,2,1}(\cdot) & \cdots & \phi_{l-1,2,n_{l-1}}(\cdot) \\
\vdots & \ddots & \vdots \\
\phi_{l-1,n_l,1}(\cdot) & \cdots & \phi_{l-1,n_l,n_{l-1}}(\cdot)
\end{pmatrix}
}_{\Phi_{l-1}}
\mathbf{h}^{l-1},
\label{ep08}
\end{align}
where $n_l$ and $n_{l-1}$ are the $l^{th}$ and $(l-1)^{th}$ hidden layer sizes. The function $\phi(x)$ is defined as:
\begin{align}
\phi(x)
= W_{b}^{l-1}\,\frac{x}{1+e^{-x}}
\;+\;
W_{s}^{l-1}\sum_i c_i\,B_i(x),
\label{ep09}
\end{align}
where $W_{b}^{l-1}$ is the base weight and $W_{s}^{l-1}$ is the spline weight for the $(l-1)^{th}$h hidden layer, $c_i$ is the control coefficient, and $B_i$ is denoted as B-splines.

Finally, the output is obtained by reweighting each channel of the original feature map with the computed channel attention coefficients, $F''_k$ is denoted as:
\begin{align}
\label{ep10}
F''_k = M_{k,c} \odot F'_k,
\end{align}

\begin{table*}[t!]
\centering
\caption{Demographic Information of Subjects Used in the Study from the Public ADNI Dataset}
\label{tbl}
\begin{tabular}{%
    >{\raggedright\arraybackslash}p{2cm}  
    >{\centering\arraybackslash}p{3.2cm}  
    >{\centering\arraybackslash}p{3.2cm}    
    >{\centering\arraybackslash}p{3.5cm}    
    >{\centering\arraybackslash}p{3.2cm}    
}
\toprule
\textbf{Category} & \textbf{Gender (Male/Female)} & \textbf{Age (Mean ± SD)} & \textbf{Education (Mean ± SD)} &\textbf{ MMSE (Mean ± SD)} \\
\midrule
CN   & 191/227 & 74.31~$\pm$~6.99 & 16.44~$\pm$~2.61 & 28.96~$\pm$~1.22 \\
sMCI  & 176/117 & 72.62~$\pm$~7.37 & 16.00~$\pm$~2.76 & 27.20~$\pm$~1.59 \\
pMCI  & 159/119 & 74.81~$\pm$~7.09 & 15.87~$\pm$~2.77 & 25.74~$\pm$~2.54 \\
AD   & 205/169 & 75.70~$\pm$~8.04 & 15.16~$\pm$~2.92 & 21.03~$\pm$~4.62 \\
\bottomrule
\end{tabular}
\end{table*}

\subsection{Hybrid Loss Function}
In this study, model training involves two classification tasks. Therefore, we employ the standard cross-entropy (CE) loss function to evaluate the effectiveness of the classification process, ${L}_{\rm CE}$ is defined as:
\begin{align}\label{ep11}
L_{\rm CE}
= -\frac{1}{M}\sum_{m=1}^{M}\bigl[(1-y'_m)\log(1-y_m)+y'_m\log(y_m)\bigr],
\end{align}
where $M$ denotes the number of subjects, $y'_m\in\{0,1\}$ represents the ground-truth label, and $y_m\in[0,1]$ indicates the predicted probability output by the model for the $m^{th}$ subject. 

To more efficiently learn disease-related discriminative features, we introduce a smoothed logits-constraint (SLC) loss function, which is combined with the CE loss to form a hybrid objective function. This approach is inspired by the logits-constraint (LC) loss proposed by Xu et al. \cite{xu2024interpretable}. In their original implementation, the LC loss was added to the CE loss in a single step after 20 epochs of training, which can disrupt model convergence and limit its effectiveness. Based on our ablation study, we find that smoothly incorporating the LC loss after 20 epochs by evenly distributing its introduction across 20 incremental steps leads to improved model stability and overall performance.

Specifically, the KANSC module produces three feature maps, $F''_a$, $F''_c$, and $F''_s$, which are combined by element-wise addition and then reshaped into $F_{total} = [F_{total,1}, F_{total,2},..., F_{total, N}] \in \mathbb{R}^{C' \times N}$, where $C'$ denotes the number of channels, $N = D' \times H' \times W'$ represents the total number of 3D patches, and $F_{total,i}\in \mathbb{R}^{C'}$ is the feature embedding of the $i^{th}$ patch. For each patch, we compute the global attention weight $M_i$ as follows:
\begin{align}\label{ep12}
M_i = \frac{1}{1+\exp\bigl(-(\mathbf{v}^\top\mathrm{ReLU}(\mathbf{w}\,F_{total,i}))\bigr)},
\end{align}
where $\mathbf{w} \in \mathbb{R}^{d \times C'}$, $\mathbf{v} \in \mathbb{R}^{d \times 1}$ are the parameters of the two fully connected layers with hidden dimension $d$. We then reweight the original features, $\widetilde F_i$ is denoted as:
\begin{align}\label{ep13}
\widetilde F_i = M_i \odot F_{total,i},
\end{align}

Subsequently, each weighted feature vector $\widetilde F_i$ is passed through the classification layer to produce position-level logits vector $\alpha_i$, which is denoted as: 
\begin{align}\label{ep14}
\alpha_i= \widetilde F_i\psi,
\end{align}
where $\psi \in \mathbb{R}^{C' \times 2}$ is the parameters of the classification layer, with 2 indicating the number of classes. Then, we compute the SLC loss, which is denoted as:
\begin{align}\label{ep15}
L_{\mathrm{SLC}}
= \sqrt{\sum_{i=1}^{T}
\Bigl(
M_{i}
- \frac{\exp(\alpha_{i,j})}
       {\exp(\alpha_{i,0})+\exp(\alpha_{i,1})}
\Bigr)^{2}
},
\end{align}
where $\alpha_{i,0}$ and $\alpha_{i,1}$ denote the model's raw scores for class 0 and class 1 at spatial position $i^{th}$, and $\alpha_{i,j}$ corresponds to the score for the true class label $j \in \{0,1\}$. Finally, the overall loss function $L_{total}$ is denoted as: 
\begin{align}\label{ep16}
L_{total} = L_{\rm CE} + \lambda L_{\rm SLC},
\end{align}
where $\lambda$ is a hyperparameter, established at 0.2 in the study.

\section{Experiment}
\label{experiment}
\subsection{Datasets and Data Preprocessing}
In our study, we employ the Alzheimer’s Disease Neuroimaging Initiative (ADNI) dataset, which is publicly accessible at \url{http://adni.loni.usc.edu}.le\ref{tbl} summarizes the demographic and clinical characteristics of all subjects, including gender, age,  education, and Mini-Mental State Examination (MMSE) score. Subjcets are classified into four groups: CN, sMCI, pMCI, and AD. Specifically, pMCI refers to MCI subjects who convert to AD within 36 months from baseline, while the remaining MCI subjects are categorized as sMCI.

All raw sMRI scans are preprocessed using a standardized pipeline, which includes format conversion (MRIcron), neck removal (FSL v6.0.7 robustfov), reorientation to standard space (FSL v6.0.7 fslreorient2std), brain extraction (HD-BET \cite{isensee2019automated}), linear registration (FSL v6.0.7 FLIRT), and bias field correction (FSL v6.0.7 FAST), with all tools applied using default parameters. After preprocessing, each sMRI is resampled to $182 \times 218 \times 182$ and then cropped to $160 \times 192 \times 160$ to remove background regions without informative brain tissue. This preprocessing yields high-quality, anatomically aligned images, which facilitates more accurate extraction of disease-related features and enhances model performance.

\subsection{Experimental Settings}
\label{Exp:set}
Our model is implemented in PyTorch and all experiments are conducted on an NVIDIA L40 GPU. To address the limited number of 3D sMRI samples, we apply data augmentation during training, including random translation (with uniform probability in all directions) and random flipping with a probability of 0.5. Model training is performed for 100 epochs with a batch size of 3, using stochastic gradient descent (SGD). For the CN versus AD classification task, the initial learning rate is set to 0.005 and reduced by a factor of 0.5 every 20 epochs. For the sMCI versus pMCI task, we employ a transfer learning strategy: we initialize with the best-performing parameters from the five CN versus AD models and set the initial learning rate to 0.001, while keeping all other hyperparameters the same. In this process, the five pretrained weight sets obtained from the CN versus AD task are individually evaluated to identify the one that achieves the highest performance. The detailed comparison results are provided in Supplementary Material Section 1. For the KAN modules, we configure the hidden layers to have 256, 32, and 256 units, and set the internal grid\_size parameter to 8, while all other parameters follow the default settings of efficientKAN.

To achieve a fair and reliable performance assessment, a five-fold cross-validation strategy is applied to both classification tasks. For each task, the dataset is randomly partitioned into five subsets, with four used for training and the remaining one for testing. This process is repeated five times, and the average performance across all folds is reported. Model performance is comprehensively assessed using widely recognized metrics, including accuracy (ACC), sensitivity (SEN), specificity (SPE), balanced F1-score (F1), and the area under the receiver operating characteristic curve (AUC).

\begin{table*}[htbp]
\centering
\caption{Performance comparison between the proposed method and seven deep learning approaches for the classification tasks of CN vs. AD and pMCI vs. sMCI}
\renewcommand{\arraystretch}{1.2}
\label{tb2}
\begin{tabular*}{\textwidth}{@{\extracolsep{\fill}}lcccccccccc}
\toprule
\multirow{3}{*}{\textbf{Method}} 
    & \multicolumn{5}{c}{\textbf{CN vs. AD}} 
    & \multicolumn{5}{c}{\textbf{sMCI vs. pMCI}} \\
\cmidrule(lr){2-6} \cmidrule(lr){7-11}
    & \textbf{Subjects} & \textbf{ACC} & \textbf{SEN} & \textbf{SPE} & \textbf{AUC} 
    & \textbf{Subjects} & \textbf{ACC} & \textbf{SEN} & \textbf{SPE} & \textbf{AUC} \\
\midrule
DCN            & 427/352 & 0.854 & 0.801 & 0.895 & 0.902 & 342/234 & 0.639 & 0.631 & 0.645 & 0.666 \\
3D-ResAttNet   & 650/353 & 0.913 & 0.910 & 0.919 & \textbf{0.984} & 232/172 & \underline{0.821} & 0.812 & 0.809 & \textbf{0.920} \\
AD²A           & 765/427 & 0.925 & 0.750 & \textbf{0.957} & 0.957 & 400/253 & 0.780 & 0.534 & \underline{0.866} & 0.788 \\
pABN           & 394/348 & 0.907 & 0.888 & 0.924 & 0.936 & 401/197 & 0.793 & 0.546 & 0.841 & 0.776 \\
MSACNet        & 427/352 & 0.893 & 0.827 & 0.945 & 0.932 & 342/234 & 0.725 & 0.606 & 0.795 & 0.721 \\
MTSSL            & 767/420 & 0.924 & 0.843 & \textbf{0.957} & 0.954 & 488/284 & 0.771 & 0.358 & \textbf{0.910} & 0.754 \\
LA-GMF         & 418/374 & \underline{0.932} & \underline{0.923} & 0.938 & 0.965 & 293/278 & 0.800 & \underline{0.820} & 0.782 & 0.846 \\
\textbf{MPF-KANSC}      & 418/374 & \textbf{0.943} & \textbf{0.931} & \underline{0.955} & \underline{0.969} & 293/278 & \textbf{0.842} & \textbf{0.835} & 0.850 & \underline{0.891} \\
\bottomrule
\multicolumn{11}{l}{The best and second-best results are shown in bold and underlined, respectively.}
\end{tabular*}
\end{table*}

\subsection{Comparison Methods}
For a fair and unbiased evaluation, we compare the proposed MPF-KANSC model with several end-to-end methods that utilize the same dataset, have a comparable number of subjects, and take the entire sMRI as input. Notably, all compared approaches employ the same diagnostic criterion, defining pMCI as MCI cases progressing to AD within 36 months, and sMCI as those without progression. A brief description of these competing methods is provided below:
\begin{enumerate}
\item DCN \cite{gao2021task}: which employs a deep convolutional network for disease classification by combining convolution and pooling layers with fully connected modules and transition layers. 
\item 3D-ResAttNet \cite{zhang2021explainable}: which extends a 3D ResNet backbone by integrating residual learning with self‑attention modules to fully leverage local and global MRI information and prevent feature loss. 
\item AD²A \cite{guan2021multi}: which proposes an attention-guided deep domain adaptation framework, integrating an sMRI encoder, label‑free attention discovery, and adversarial alignment to harmonize source and target sMRI domains. 
\item pABN \cite{guan2022parallel}: which employs a 3D CNN backbone with parallel attention‐augmented bilinear modules to extract global and local sMRI features directly, without any handcrafted priors. 
\item MSACNet \cite{gao2023hybrid}: which utilizes a multi-scale attention convolutional network with dense connections to adaptively capture and fuse rich, discriminative features from the entire brain sMRI. 
\item MTSSL \cite{han2023multi}: which employs a Multi-Template Self-Supervised Learning framework by using a simple Siamese network, disentangling class-related embeddings through cosine similarity learning.
\item LA-GMF \cite{xu2024interpretable}: which leverages logits-constraint attention and graph-based multi-scale fusion to model sMRI data, capturing patch interactions and feature complementarity for superior feature representations. 
\end{enumerate}

Among the compared methods, 3D-ResAttNet, AD²A, and pABN are implemented with only sMRI inputs. For DCN, MSACNet, and MTSSL, although their original studies included additional clinical data, we report the results of the corresponding submodels that use only sMRI data to ensure a fair comparison. Additionally, since the code for LA-GMF is available, we reproduced this method on our dataset, and the results shown in Table \ref{tb2} are based on our implementation.

\subsection{Classification Results}
Table \ref{tb2} presents the comparative performance of all methods across four evaluation metrics on the CN vs. AD and sMCI vs. pMCI classification tasks. The proposed MPF-KANSC consistently achieves the highest overall accuracy, reaching 94.3\% in the CN vs. AD task and 84.2\% in the sMCI vs. pMCI task. It also obtains the best sensitivity, with values of 93.1\% and 83.5\% for the two tasks, respectively. In terms of specificity and AUC, our model also achieves competitive results compared to existing methods. Overall, these results highlight the strong generalizability and clinical potential of MPF-KANSC, which maintains stable and reliable detection of disease-related cases across different classification scenarios.

Such competitive performance is mainly attributed to three key aspects of our model's design. Firstly, the KANSC attention mechanism enables effective encoding of high-dimensional sMRI data, allowing the model to extract more informative and discriminative semantic features. Secondly, by incorporating pathological information from three anatomical planes, the model captures richer spatial structural representations, which are often overlooked in single-view approaches. Thirdly, the use of a hybrid loss module facilitates a better integration of global information and explicitly aligns attention distributions with semantic categories. These components work synergistically to enhance the model’s representational capacity and contribute to its competitive predictive performance.

\begin{figure*}[t]
	\centering
	\includegraphics[width=\textwidth]{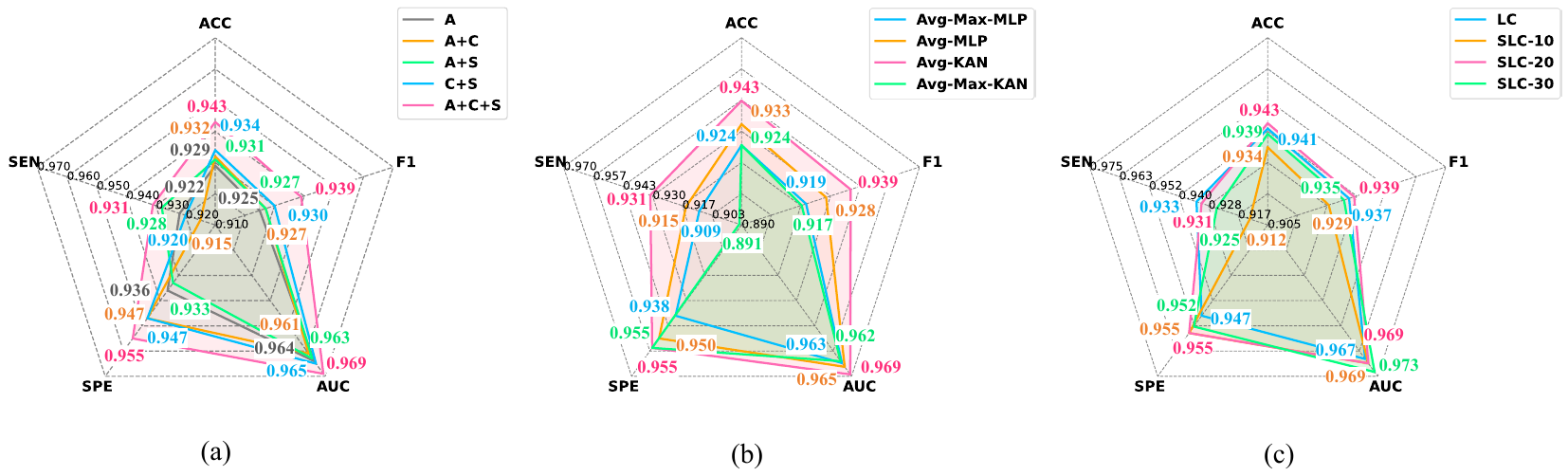}
	\caption{Ablation results for different internal parameter settings of the three modules on the CN vs. AD task. (a) Results for different combinations of planes: gray for Axial, orange for Axial and Coronal, green for Axial and Sagittal, blue for Coronal and Sagittal, and pink for three planes. (b) Configurations of the spatial-channel attention mechanism: blue for the traditional structure (Maxpool, Avgpool, and MLP in the channel module); orange for Avgpool and MLP; pink for Avgpool and KAN; green for Maxpool, Avgpool, and KAN. (c) Strategies for introducing the SLC loss: blue for adding the LC loss all at once after 20 epochs; orange, pink, and green for gradually introducing the SLC loss in 10, 20, and 30 steps after 20 epochs, respectively. The annotated values in (a), (b), and (c) range from 0.910–0.970, 0.890–0.970, and 0.905–0.975, respectively.}
	\label{FIG:04}
\end{figure*}

\begin{table*}[htbp]
\centering
\caption{Ablation study results for both the CN vs. AD and sMCI vs. pMCI classification tasks}
\renewcommand{\arraystretch}{1.2}
\label{ablation}
\begin{tabular*}{\textwidth}{@{\extracolsep{\fill}}lcccccccccc}
\toprule
\multirow{3}{*}{\textbf{Method}} 
    & \multicolumn{5}{c}{\textbf{CN vs. AD}} 
    & \multicolumn{5}{c}{\textbf{sMCI vs. pMCI}} \\
\cmidrule(lr){2-6} \cmidrule(lr){7-11}
    & \textbf{ACC} & \textbf{SEN} & \textbf{SPE} & \textbf{AUC} & \textbf{F1}
    & \textbf{ACC} & \textbf{SEN} & \textbf{SPE} & \textbf{AUC} & \textbf{F1}
    \\
\midrule
Baseline                  & 0.908   & 0.877   & 0.935   & 0.952   & 0.901  & 0.800 & 0.766 & 0.833 & 0.856 & 0.787   \\
Baseline+KANSC            & 0.921   & 0.904   & 0.935   & 0.957   & 0.915   & 0.807 & \underline{0.827} & 0.789 & 0.856 & \underline{0.808}   \\
Baseline+KANSC+SLC        & \underline{0.929}   & \underline{0.923}  & \underline{0.936}   & \underline{0.964}   & \underline{0.925}   & \underline{0.809} & 0.777 & \underline{0.840} & \underline{0.865} & 0.797 \\
\textbf{MPF-KANSC}             & \textbf{0.943}   & \textbf{0.931}   & \textbf{0.955}   & \textbf{0.969}   & \textbf{0.939}   & \textbf{0.842} & \textbf{0.835} & \textbf{0.850} & \textbf{0.891} & \textbf{0.837}
\\
\bottomrule
\end{tabular*}
\end{table*}

\subsection{Ablation Studies}
\label{ablation studies}
We perform ablation studies to systematically assess the contributions of key components in our framework. Specifically, for both the CN vs. AD and sMCI vs. pMCI classification tasks, we evaluate four methods: Baseline, Baseline+KANSC, Baseline+KANSC+SLC, and MPF-KANSC. The Baseline utilizes a single-plane CNN module based on axial features. Baseline+KANSC incorporates the KANSC attention mechanism, Baseline+KANSC+SLC further adds the smoothed logits-constraint loss, and MPF-KANSC denotes the complete model with multi-plane fusion. All other experimental settings are kept consistent with those described in Section \ref{Exp:set}, except for the components being ablated.

As shown in Table \ref{ablation}, each component incrementally contributes to the overall performance improvement across both classification tasks. For the CN vs. AD task, introducing the KANSC attention module raises the accuracy from 90.8\% to 92.1\% and the F1-score from 90.1\% to 91.5\%. Incorporating the SLC loss further increases accuracy and the F1-score to 92.9\% and 92.5\%, respectively. The full MPF-KANSC model with multi-plane fusion achieves the best results, with an accuracy of 94.3\% and an F1-score of 93.9\%. Sensitivity, specificity, and AUC similarly improve, reaching 93.1\%, 95.5\%, and 96.9\%, respectively. For the sMCI vs. pMCI classification task, the incremental gains in accuracy from adding the KANSC and SLC modules are relatively modest, which may be attributed to the strong baseline established through transfer learning. However, introducing multi-plane fusion in the full MPF-KANSC model results in notable improvements across all evaluation metrics. In particular, sensitivity increases from 76.6\% in the baseline to 83.5\%, specificity rises from 83.3\% to 85.0\%, AUC improves from 85.6\% to 89.1\%, and the F1-score grows from 78.7\% to 83.7\%. These findings indicate that, despite a robust baseline, the addition of attention and loss modules, especially when combined with multi-plane fusion, brings substantial benefits for capturing complex disease patterns and further enhances diagnostic robustness.

Furthermore, we conduct additional ablation studies on the CN vs. AD classification task to evaluate the impact of key internal parameters. Specifically, we investigate the effectiveness of multi-plane fusion, various configurations of the spatial-channel attention mechanism, and different scheduling strategies for the SLC loss, as illustrated in Fig. \ref{FIG:04}. The corresponding ablation experiments for the sMCI vs. pMCI classification task are provided in Supplementary Material Section 2.

From Fig. \ref{FIG:04}(a),  we can observe that incorporating any two anatomical planes consistently improves performance over single-plane models, highlighting the benefits of multi-view integration. Fusion of all three anatomical planes achieves the highest scores across all five evaluation metrics. Specifically, compared to the single-plane (axial) model, three-plane fusion improves accuracy by 1.4\%, sensitivity by 0.9\%, specificity by 1.9\%, AUC by 0.5\%, and F1-score by 1.4\%. The results suggest that multi-plane feature fusion significantly strengthens the model’s capability to extract diagnostically relevant features, enabling more comprehensive spatial information extraction from sMRI and supporting more accurate disease identification.

As illustrated in Fig. \ref{FIG:04}(b), the spatial-channel configurations that use only Avgpool in the channel module (pink and orange lines) demonstrate superior performance compared to those employing mixed Maxpool and Avgpool strategies (blue and green lines). This suggests that, in our model, the use of Maxpool causes certain pathological features to be overlooked by focusing only on the largest responses \cite{zafar2022comparison}. Therefore, employing only Avgpool achieves better training outcomes. Notably, compared to the Avg-MLP architecture, our proposed Avg-KAN model achieves substantial improvements in all metrics, with accuracy increasing by 1.0\% and the F1-score by 1.1\%. This underscores the unique multivariate function decomposition capability of KAN, which enables more effective modeling of the complex relationships within sMRI data. Moreover, as illustrated in Fig. \ref{FIG:04}(c), applying SLC with a smoothing step of 20 achieves the most stable and robust classification results, indicating that the gradual introduction of logits-constraint loss stabilizes optimization and enhances overall model performance. Collectively, these findings further substantiate the effectiveness and reliability of our proposed framework and its key components.

\begin{figure*}[t]
	\centering
	\includegraphics[width=\textwidth]{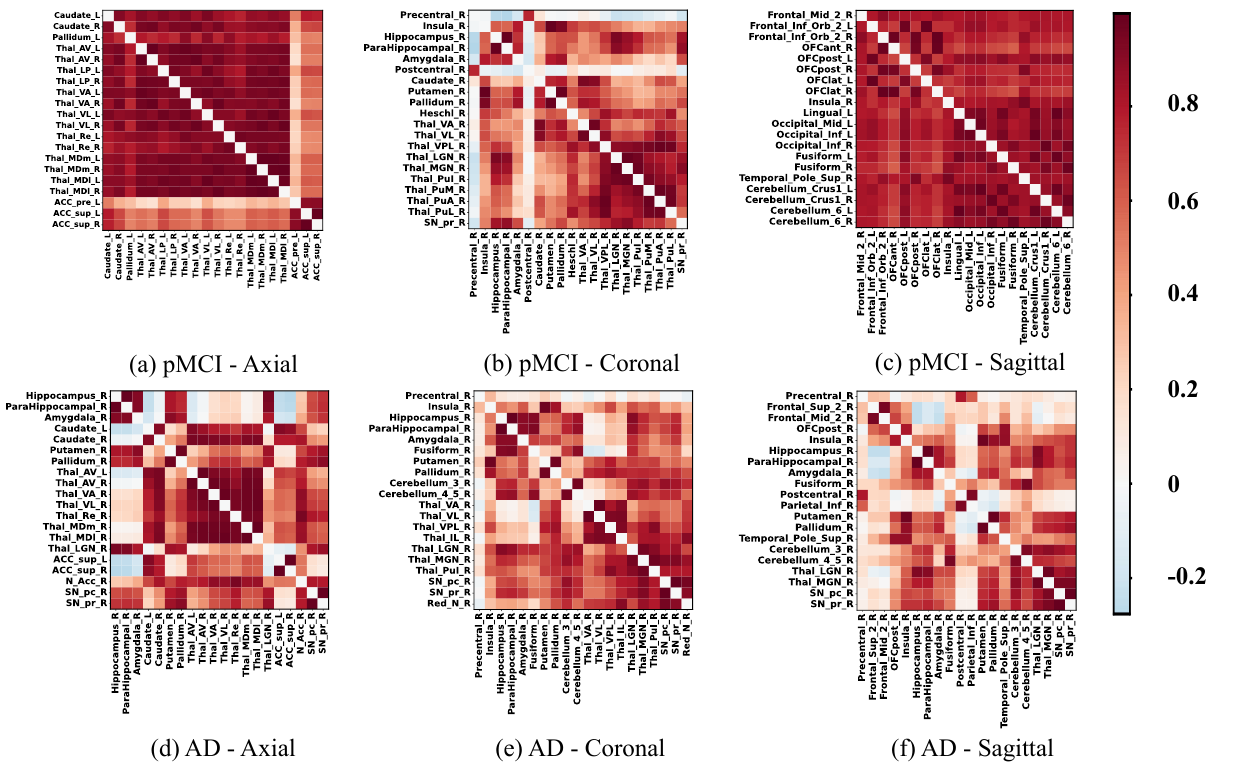}
	\caption{Heatmaps of correlation matrices among the 20 most important brain regions in pMCI and AD groups for two classification tasks. (a)-(c) The correlation matrices for pMCI subjects learned from the axial, coronal, and sagittal planes, respectively. (d)-(f) The correlation matrices for AD subjects. The color of each edge indicates the strength of correlation between regions.}
	\label{FIG:05}
\end{figure*}

\begin{table*}[t]
\centering
\caption{Top 20 brain regions predicted by the model for each plane and two classification tasks}
\renewcommand{\arraystretch}{1.2}
\label{top 20}
\begin{tabular}{llp{13cm}}
\toprule
\textbf{Experiment} & \textbf{Plane} & \multicolumn{1}{c}{\textbf{Top 20 Prediction Regions}} \\
\midrule
\multirow{3}{*}{\textbf{CN vs. AD}}
& Axial    & 46, 80, 122, 126, 155, 42, 121, 134, 164, 128, 78, 156, 158, 138, 76, 162, 44, 75, 136, 140 \\
& Coronal  & 46, 42, 44, 140, 80, 78, 142, 100, 34, 164, 144, 162, 130, 102, 2, 128, 126, 132, 60, 166 \\
& Sagittal & 46, 42, 44, 78, 30, 2, 34, 80, 60, 140, 62, 66, 88, 164, 102, 6, 100, 4, 142, 162 \\
\midrule
\multirow{3}{*}{\textbf{sMCI vs. pMCI}}
& Axial    & 155, 121, 122, 75, 156, 125, 124, 123, 126, 135, 137, 136, 127, 138, 76, 128, 133, 134, 153, 79 \\
& Coronal  & 46, 42, 78, 80, 34, 140, 2, 44, 84, 130, 144, 142, 150, 128, 126, 62, 164, 148, 146, 76 \\
& Sagittal & 57, 32, 58, 12, 55, 30, 31, 59, 88, 103, 104, 6, 95, 28, 29, 34, 96, 60, 51, 11 \\
\bottomrule
\multicolumn{3}{p{\textwidth}}{The numbers refer to region indices defined by the brain automatic anatomical labeling AAL 3.}
\end{tabular}
\end{table*}

\subsection{Interpretable Studies}
In this section, we apply the Gradient-weighted Class Activation Mapping (Grad-CAM) \cite{selvaraju2017grad} to obtain the average Grad-CAM values across the three anatomical planes for both classification tasks. The information from the coronal and axial planes is realigned to the axial orientation to ensure spatial correspondence with the AAL3 template \cite{rolls2020automated}, after which the activation maps are segmented into distinct brain regions. The corresponding brain region indices in AAL3 are listed in Supplementary Material Section 3. As shown in Table \ref{top 20}, we list the top 20 predicted brain regions across the three planes for both classification tasks. For AD classification, the top five regions are Amygdala\_R \cite{poulin2011amygdala}, Hippocampus\_R \cite{rao2022hippocampus}, ParaHippocampal\_R \cite{van2000parahippocampal}, Pallidum\_R \cite{iizuka2023impaired}, and Thal\_AV\_R \cite{adebisi2021differential}. For the MCI conversion prediction task, the top five regions are Amygdala\_R \cite{qu2023volume}, Hippocampus\_R \cite{qu2023volume}, Putamen\_R \cite{farid2014ic}, Pallidum\_R \cite{tang2014shape}, and ACC\_sup\_L \cite{shukla2020quantitation}. These regions are primarily associated with emotional regulation, learning, memory, and motor functions, and dysfunction in these domains is frequently observed throughout the progression of AD.

\begin{figure*}
	\centering
	\includegraphics[width=\textwidth]{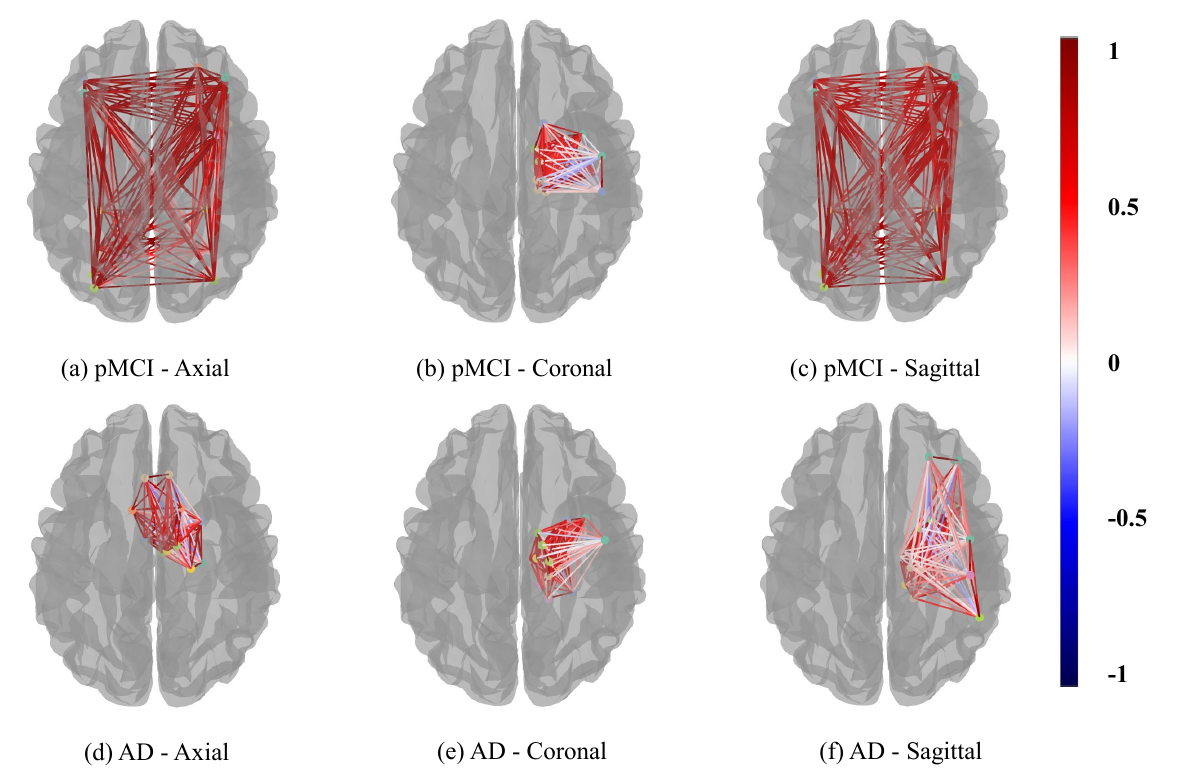}
	\caption{Visualization of the spatial connectivity among the top 20 most important brain regions for the pMCI and AD groups. (a)-(c) The connectivity graphs for pMCI, derived from the axial, coronal, and sagittal planes, respectively. (d)-(f) The corresponding results for AD. All graphs are mapped onto the same brain template for consistent comparison and clearer spatial presentation. The color of each edge represents the strength of correlation between regions.}
	\label{FIG:06}
\end{figure*}

To further explore how brain region associations evolve during disease progression, we compute correlation matrices of the Grad-CAM-derived values for the top 20 most important regions in both classification tasks and visualize them as heatmaps (see Fig. \ref{FIG:05}). Here, we focus on the pMCI and AD groups to better illustrate the changes related to disease advancement. For pMCI patients, the correlation matrices show strong inter-regional associations in the axial and sagittal planes, while the coronal plane displays relatively weaker connections. This indicates that, at the early stage of the disease, the model highlights coordinated patterns across major brain regions. In contrast, for AD patients, the overall correlations among these key regions are much weaker in all three planes, as reflected by the lighter colors in the heatmaps. These results suggest that, from the model's perspective, the attention patterns across different regions become less coordinated as the disease progresses. The reduced coordination among brain regions, as revealed by the model, exhibits a pattern similar to that reported in studies on network disruption in AD \cite{dillen2017functional}, \cite{huang2018characteristic}, \cite{sun2024alzheimer}.

Moreover, we further visualize the spatial distribution of the most critical brain regions to examine hemispheric asymmetry in AD progression, as shown in Fig. \ref{FIG:06}. Notably, we observe that important regions identified in AD patients predominantly localized within the right hemisphere. Interestingly, the pMCI patients already exhibit this emerging rightward dominance in the coronal plane, suggesting that hemispheric asymmetry in subcortical structures may appear earlier than previously assumed. Traditionally, AD-related hemispheric asymmetry has primarily been described based on cortical thinning of gray matter \cite{derflinger2011grey}, \cite{kim2012cortical}, typically greater in the left hemisphere. However, recent evidence highlights subcortical structures such as the hippocampus and amygdala as potentially more sensitive markers of disease progression. For instance, Shi et al. \cite{shi2009hippocampal} observed consistent asymmetry in hippocampal volume loss across CN, MCI, and AD groups, with the left hippocampus smaller than the right. Adebisi et al. \cite{adebisi2021differential} identified the Thal-AV-R region as critical for distinguishing AD from CN networks. Qu et al. \cite{qu2023volume} further reported that atrophy in the hippocampus and amygdala predominantly affects the right hemisphere. Abdelaziz et al. \cite{abdelaziz2025multi} also reported a right hemisphere bias in regions distinguishing AD from CN but noted more bilateral distribution in regions distinguishing sMCI from pMCI, which aligns well with our findings. However, leveraging a multi-plane learning strategy, our model uniquely identifies this rightward asymmetry earlier in disease progression, enhancing sensitivity to disease advancement. Overall, our approach contributes novel evidence supporting subcortical hemispheric asymmetry as an important feature of AD progression.

\section{Conclusion}
\label{conclusion}
In this study, we propose MPF-KANSC, a novel interpretable deep learning framework that integrates multi-plane fusion and a KAN-guided spatial-channel attention mechanism for AD diagnosis and MCI prediction using sMRI. Our approach independently models spatial and channel features from the axial, coronal, and sagittal planes via 3D CNNs and KANSC attention, enabling comprehensive representation of structural information from each anatomical perspective. By fusing these features at the output stage and incorporating a smoothed logits-constraint loss, MPF-KANSC achieves superior performance on both CN vs. AD and sMCI vs. pMCI classification tasks, with accuracies reaching 94.3\% and 84.2\%, respectively. Importantly, our model demonstrates superior interpretability, as experimental results reveal that right-hemisphere subcortical regions play an increasingly critical role in disease discrimination during AD progression. This finding provides new evidence for asymmetric structural changes, thereby advancing our understanding of AD pathology. 

In future work, we will extend our framework to incorporate multimodal clinical data and longitudinal analysis, further enhancing its generalizability and utility in real-world clinical applications.

\section*{References}
\bibliographystyle{IEEEtran}
\bibliography{cas-refs} 
\end{document}